\pdfoutput=1

\documentclass[11pt]{article}

\usepackage{ACL2023}

\usepackage{times}
\usepackage{latexsym}
\usepackage{subcaption}
\usepackage{graphicx}
\usepackage{amsmath} 
\usepackage{booktabs}
\usepackage{multirow} 
\usepackage{paralist, tabularx}

\usepackage{amsfonts}
\usepackage{adjustbox}
\usepackage{color,colortbl}
\usepackage{xcolor,xspace}
\usepackage{algorithm}
\usepackage{algpseudocode}
\usepackage{xtab}
\usepackage{multicol}
\usepackage{makecell}

\newtheorem{definition}{Definition}

\usepackage[T1]{fontenc}

\usepackage[utf8]{inputenc}

\usepackage{microtype}

\usepackage{inconsolata}

\newcommand*{\Scale}[2][4]{\scalebox{#1}{$#2$}}

\title{CodeTaxo: Enhancing Taxonomy Expansion with Limited Examples via Code Language Prompts}

\author{Qingkai Zeng\textsuperscript{\rm 1}\footnotemark[1], Yuyang Bai\textsuperscript{\rm 1}, Zhaoxuan Tan\textsuperscript{\rm 1}, Zhenyu Wu\textsuperscript{\rm 1}, Shangbin Feng\textsuperscript{\rm 2}, Meng Jiang\textsuperscript{\rm 1} \\
        \textsuperscript{\rm 1}University of Notre Dame
        \textsuperscript{\rm 2} University of Washington \\
        \{qzeng, ybai3, ztan3, zwu23, mjiang2\}@nd.edu,  shangbin@cs.washington.edu}

\begin{document}
\maketitle

\renewcommand{\thefootnote}{\fnsymbol{footnote}}
\footnotetext[1]{Corresponding author.}
\renewcommand{\thefootnote}{\arabic{footnote}}

\begin{abstract}
Taxonomies provide structural representations of knowledge and are crucial in various applications.  The task of taxonomy expansion involves integrating emerging entities into existing taxonomies by identifying appropriate parent entities for these new query entities. Previous methods rely on self-supervised techniques that generate annotation data from existing taxonomies but are less effective with small taxonomies (fewer than 100 entities).  In this work, we introduce \textsc{CodeTaxo}, a novel approach that leverages large language models through code language prompts to capture the taxonomic structure. Extensive experiments on five real-world benchmarks from different domains demonstrate that \textsc{CodeTaxo} consistently achieves superior performance across all evaluation metrics, significantly outperforming previous state-of-the-art methods. The code and data are available at \url{https://github.com/QingkaiZeng/CodeTaxo-official}.
\end{abstract}

\section{Introduction}

Taxonomies are hierarchical structures encoding hypernym–hyponym (i.e., “is-A”) relations between concepts or entities. Relational knowledge derived from taxonomies has been widely leveraged to identify semantic relevance for web search~\cite{yin2010building,liu2020giant,kang2024improving}, personalized recommendation~\cite{zhang2014taxonomy,tan2022enhancing,huang2019taxonomy}, and question answering~\cite{yang2017efficiently}. However, existing taxonomies are mainly constructed by experts or through crowd-sourcing, making the process time-consuming, labor-intensive, and restricted in coverage~\cite{bordea2016semeval,jurgens2016semeval}. As new entities emerge, continually enriching taxonomies with these additions becomes vital. To address these challenges, taxonomy expansion aims to integrate new entities into existing taxonomies automatically.

\begin{figure}[t]
\centering
\begin{subfigure}[b]{0.45\textwidth}
   \includegraphics[width=\textwidth]{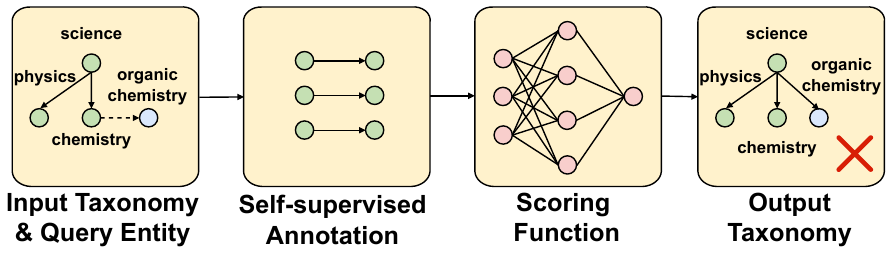}
   \caption{\textbf{Discriminative Methods:} A trained scoring function selects the most appropriate parent entity from the taxonomy for a given query entity.}
   \label{fig:motivation_sub1}
\end{subfigure}
\begin{subfigure}[b]{0.45\textwidth}
   \includegraphics[width=\textwidth]{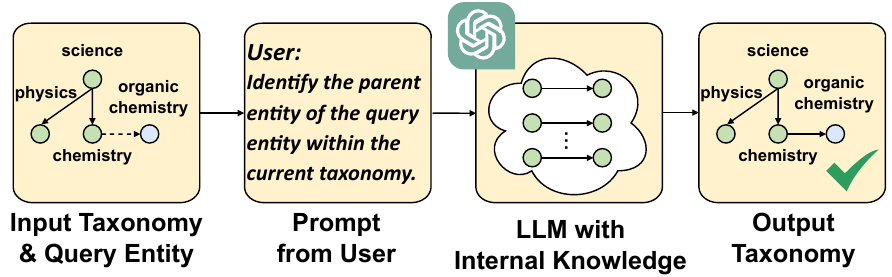}
   \caption{\textbf{Generative Methods:} LLMs generate the parent entity from the taxonomy based on the query entity.}
   \label{fig:motivation_sub2}
\end{subfigure}
\caption{Two Types Taxonomy Expansion Methods}
\vspace{-0.3in}
\label{fig:motivation}
\end{figure}

\begin{figure*}[t]
    \centering
    \includegraphics[width=1.0\linewidth]{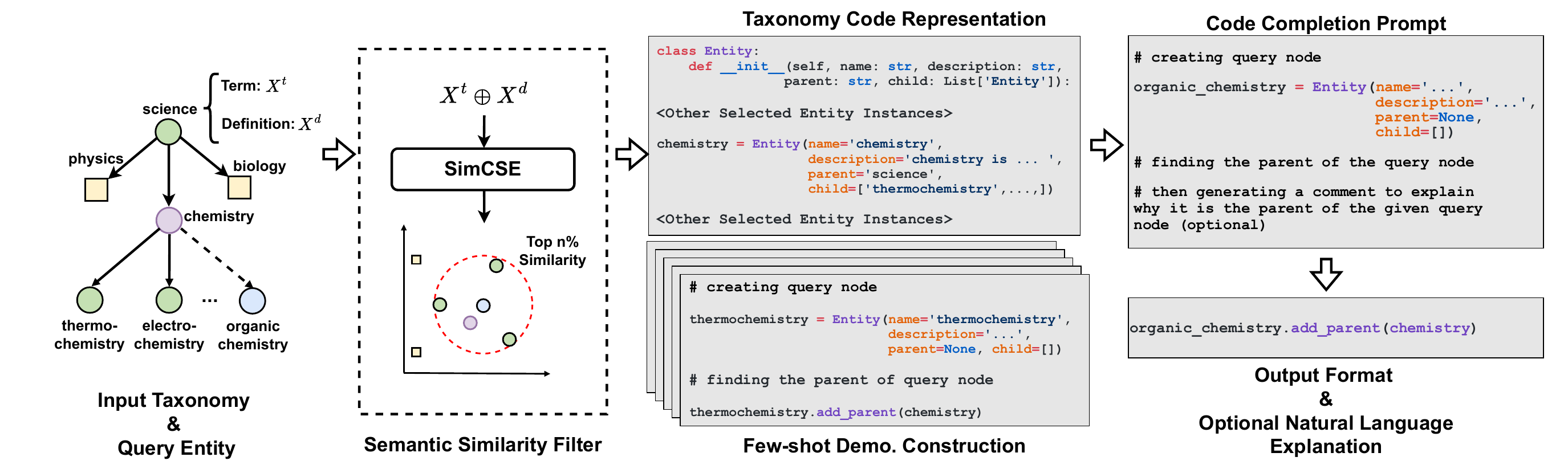}
    \vspace{-0.3in}
    \caption{The overview of the pipeline for \textsc{CodeTaxo}: \textsc{CodeTaxo} reformulates the task of integrating a query entity $q$ into an existing taxonomy $\mathcal{T}_0$ as a code completion task using code-based prompts for LLMs.}
    \label{fig:framework}
    \vspace{-0.2in}
\end{figure*}

As shown in Figure~\ref{fig:motivation_sub1}, recent taxonomy expansion methods mainly rely on discriminative methods that model hierarchical structures through techniques like Egonets~\cite{shen2020taxoexpan}, mini-paths~\cite{yu2020steam}, and Ego-Trees~\cite{wang2021enquire}. Although pre-trained language models (PLMs) enhance these methods by encoding entities’ textual descriptions~\cite{wang2021enquire,wang2022qen, liu2021temp,xu2022taxoprompt}, their reliance on limited self-supervised annotations often restricts performance. In contrast, Generative Large Language Models (LLMs) such as GPT-4~\cite{achiam2023gpt} and Llama family~\cite{touvron2023llama, dubey2024llama} have recently shown remarkable capabilities in text comprehension and generation, making them highly effective for tasks aimed at generating structural knowledge~\cite{ye2022generative, bi2024codekgc, sun2024head,sun2024large}. Increasing LLM parameters boosts generalization, surpassing smaller models and enabling superior few-shot or zero-shot performance. Even with limited annotations, LLMs effectively leverage extensive knowledge embedded within their parameters, acquired from large-scale pre-training corpora. In Figure~\ref{fig:motivation_sub2}, we illustrate the pipeline to how generative methods are applied to the taxonomy expansion task.

Two key challenges arise when applying LLMs to taxonomy expansion. First, unlike traditional text-to-text NLP tasks such as question answering and machine translation, representing the taxonomic structure for this task in natural language is inherently challenging. Specifically, the process requires "flattening" the taxonomy into a sequence of parent-child entity pairs~\cite{madaan2022language}, effectively serializing a hierarchical structure into linear text. This serialized format is notably different from the unstructured text that LLMs primarily encounter during pre-training. Furthermore, while semantically related words in natural language are usually located near each other, linearizing a taxonomy can separate conceptually related entities by significant distances within the sequence. This disparity adds to the difficulty of aligning LLM outputs with the desired structured representation. Second, scaling to large taxonomies amplifies the problem, as including every entity from the existing taxonomy in the prompt is infeasible. The limited contextual window size of current LLMs and the associated computational overhead imposes strict constraints. Even if it were possible to include thousands of entities within a prompt, the resulting structural information loss would impair the clarity of entity-specific distinctions, reducing the model's capacity to effectively utilize the taxonomy.

To overcome these challenges, we propose \textsc{CodeTaxo}, a novel taxonomy expansion approach that leverages code language as prompts. Code-based representations have shown promise in structure prediction tasks~\cite{madaan2022language, li2023codeie, wang2023code4struct,li2024knowcoder,bi2024codekgc}, as code languages provide a more natural format for structural data. In \textsc{CodeTaxo}, we frame taxonomy expansion as a code completion task. We introduce a base \texttt{Entity} class to store entity surface names, definitions, parent references, and child lists, along with two methods for modifying the taxonomic relations between entities. Each existing taxonomy entity is instantiated as a corresponding \texttt{Entity} object. Due to constraints of contextual window size, we apply a similarity-based filter, using SimCSE~\cite{gao2021simcse} to encode textual description for entities, to include only the most relevant entities in the prompt


We evaluate \textsc{CodeTaxo} through extensive experiments on two sets of small-scale WordNet and Graphine sub-taxonomies~\cite{bansal2014structured,liu2021graphine}, as well as three large-scale SemEval-2016 taxonomies~\cite{bordea2016semeval}. Our one-shot \textsc{CodeTaxo} surpasses all self-supervised baselines trained on large-scale SemEval-2016 annotations, achieving relative accuracy improvements of 10.26\%, 8.89\%, and 9.21\% on SemEval-Sci, SemEval-Env, and SemEval-Food, respectively. Additionally, we evaluated \textsc{CodeTaxo} using various open-source LLMs, revealing several interesting observations discussed in this work.

In summary, our main contributions include:
\begin{itemize}
    \item We introduce \textsc{CodeTaxo}, an innovative in-context learning method that utilizes code language prompts to represent taxonomic relationships between entities, thereby improving the effectiveness of taxonomy expansion.
    \item We develop a similarity-based filter, which employs a small pre-trained model to encode the textual descriptions of entities, ensuring that only highly relevant entities are included in the prompt concerning the query entity.
    \item Extensive experiments demonstrate that \textsc{CodeTaxo} significantly enhances the performance of taxonomy expansion across two sets of small-scale sub-taxonomies and three large-scale taxonomies.
\end{itemize}
\label{sec:intro}

\section{Problem Definition}


\begin{figure}[t]
    \centering
    \includegraphics[width=1\linewidth]{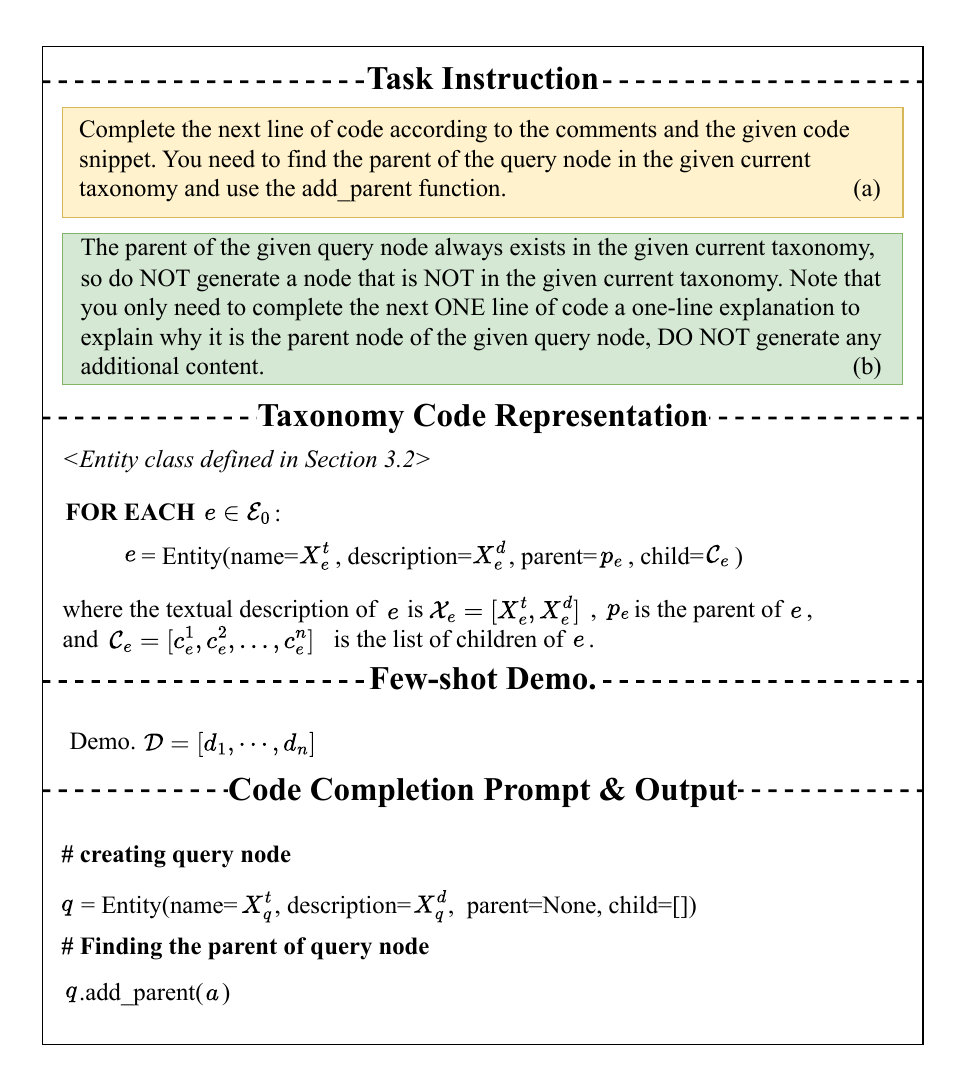}
    \vspace{-0.4in}
    \caption{Prompt Overview of \textsc{CodeTaxo}}
    \vspace{-0.2in}
    \label{fig:prompt-overview}
\end{figure}

\begin{definition}[Taxonomy]
    We follow the definition of taxonomy in~\cite{jiang2023single}.  A taxonomy $\mathcal{T} = (\mathcal{E}, \mathcal{H})$ is a tree-like structure, where each entity $e \in \mathcal{E}$ is a conceptual entity, and each edge $h \in \mathcal{H}$ represents the hypernymy-hyponymy relation between the two entities connected by it. Each entity $e$ is associated with a set of textual description $\mathcal{X}_e = \{X^t_e, X^d_e \}$, where $X^t_e$ is its term and $X^d_e$ is its definition. Meanwhile, each directed edge $h = \langle p, c \rangle \in \mathcal{H}$ represents a parent-child relationship that points to a child entity $c$ from its most exact hypernymy entity $p$. 
\end{definition}


\begin{definition}[Taxonomy Expansion]
    Given a set of emerging conceptual entities $\mathcal{E}'$, taxonomy expansion aims to incorporate these entities into an existing \textit{seed taxonomy} $\mathcal{T}_0 = (\mathcal{E}_0, \mathcal{H}_0)$. The goal is to expand $\mathcal{T}_0$ to be a larger taxonomy  $\mathcal{T} = ( \mathcal{E}_0 \cup \mathcal{E}', \mathcal{H}' )$.  To insert each query entity $q \in \mathcal{E}'$, we identify an appropriate anchor entity $a \in \mathcal{E}_0$, and introduce a new edge $\langle q, a \rangle$. Consequently, the updated edge set is $\mathcal{H}' = \mathcal{H}_0 {\cup}_{q \in \mathcal{E}'} \{\langle q, a \rangle \} $.  
\end{definition}

\section{Methodology}



In this section, we provide a comprehensive overview of our proposed \textsc{CodeTaxo} designed for addressing the taxonomy expansion task. Specifically, \textsc{CodeTaxo} expands the existing taxonomy by prompting LLMs with code language. The pipeline of \textsc{CodeTaxo} is shown in Figure~\ref{fig:framework}. Our \textsc{CodeTaxo} consists of three parts: Task Instruction, Taxonomy Code Representation, and Few-shot Demonstrations Construction.

\subsection{Task Instruction}

To enhance the effectiveness and accuracy of LLMs in completing the taxonomy expansion task, we propose a detailed task description along with a set of fundamental rules, denoted as $\mathcal{R}$, for expanding the existing taxonomy via the query entity. As illustrated in Figure~\ref{fig:prompt-overview}, component (a) outlines the objectives of the taxonomy expansion task, framing it as a code completion task and specifying \texttt{add\_parent} function should be employed. In component (b), we emphasize a set of fundamental rules $\mathcal{R}$ for the taxonomy expansion task. These rules include the following: 1. Do not use entities that are not covered in the existing taxonomy $\mathcal{T}_0 = (\mathcal{E}_0, \mathcal{H}_0)$ ($\mathbf{r}_1$); 2. Maintain the output generation format by LLMs, consisting of one line of code followed by one line explaining why the model made that prediction ($\mathbf{r}_2$); 3. Refrain from generating additional content ($\mathbf{r}_3$). Additionally, the rule for generating an explanation for the prediction in $\mathbf{r}_2$ is optional for future analysis. In \textsc{CodeTaxo}, this rule is omitted as generating explanations is not required.

\begin{figure}[t]
    \centering
    \includegraphics[width=1\linewidth]{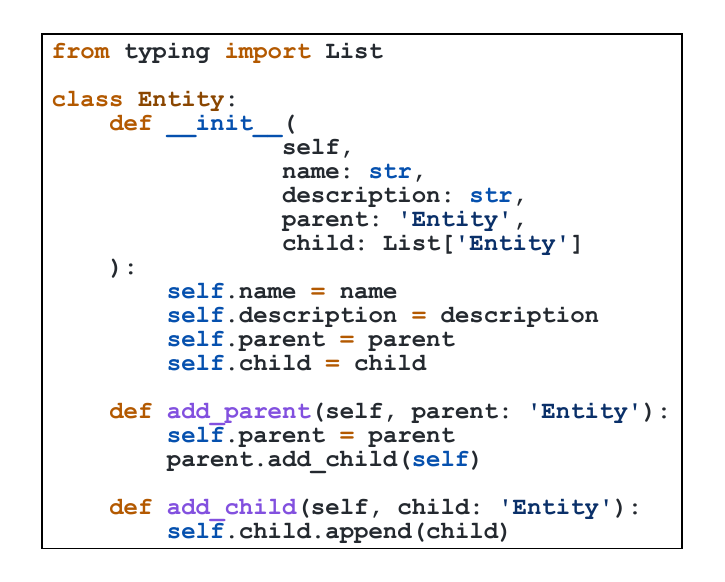}
    \vspace{-0.4in}
    \caption{Python code in \textsc{CodeTaxo} defining a \texttt{Entity} class for managing parent-child relations.}
    \label{fig:entity-representation}
    \vspace{-0.2in}
\end{figure}

\subsection{Taxonomy Code Representation}

To represent the existing taxonomy $\mathcal{T}_0 = (\mathcal{V}_0, \mathcal{E}_0)$ as code language, we concatenate the entity class definition, representation of existing taxonomic relations, and the code completion prompt. We use Python as the programming language for the code prompt due to its widespread popularity.

\subsubsection{Entity Class Definition} 
\label{sec:entity_class_definition}
First, we define a base type \texttt{Entity} to be inherited by each entity mentioned in the taxonomy expansion. In Figure~\ref{fig:entity-representation}, we define a Python class named \texttt{Entity} that models a taxonomic structure with parent-child relations. The first line imports the \texttt{List} type from the \texttt{typing} module, which is used for type hinting. This allows the \texttt{child} attribute to be explicitly declared as a list of \texttt{Entity} objects.

The \texttt{Entity} class encapsulates the attributes and methods for managing hierarchical entities. The \texttt{\_\_init\_\_} method initializes an instance of the \texttt{Entity} class with the following parameters: 

\begin{compactitem}
    \item \texttt{name}: A string storing the term of the entity.
    \item \texttt{description}: A string storing the textual description of the entity
    \item \texttt{parent}: An instance of the \texttt{Entity} class,  denoting the parent entity within the taxonomy.
    \item \texttt{child}: A list of entities, each an instance of the \texttt{Entity} class, storing the entity's children.
\end{compactitem}

These instance attributes are assigned as follows: \texttt{self.name}, \texttt{self.description}, \texttt{self.parent}, and \texttt{self.child}. Additionally, since we consider that each entity in the taxonomy should only have one parent entity, we do not use the \texttt{List} type for the \texttt{parent} attribute, unlike the \texttt{child} attribute.

The \texttt{Entity} class includes two methods for modifying the parent-child relations between entities. The first method, \texttt{add\_parent}, assigns a parent entity to the current entity. It takes one parameter, \texttt{parent}, which is an instance of the \texttt{Entity} class. The second method, \texttt{add\_child}, appends the child entity to the \texttt{self.child} list of the current entity. This method also requires one parameter, \texttt{child}, which is an instance of the \texttt{Entity} class.

\subsubsection{Representing the Existing Taxonomy}
To facilitate the taxonomy expansion, the initial taxonomy $\mathcal{T}_0$ is encoded using a programming language. Instances of the \texttt{Entity} class, as defined in Section~\ref{sec:entity_class_definition}, are created for each entity $e$ in the set $\mathcal{E}_0$ of $\mathcal{T}_0$. The taxonomy $\mathcal{T}_0$ is traversed from top to bottom, and for each entry, an entity $e \in \mathcal{E}_0$ is instantiated as follows:
\begin{align*}
    e = \operatorname{Entity}(\text{name} &= {X}^{t}_e, \text{description} ={X}^{d}_e, \\ \text{parent} &= p_e, \text{child} = \mathcal{C}_e) 
\end{align*}
where $p_e$ is the parent entity of $e$, and $\mathcal{C}_e = [c^1_e, c^2_e, \dots, c^n_e]$ is the list of its child entities.

\begin{table*}[t]
\centering
\resizebox{2\columnwidth}{!}{
\begin{tabular}{lcccccccccc}

\toprule
\textbf{Dataset} & \multicolumn{2}{c}{\textbf{SemEval-Sci}} & \multicolumn{2}{c}{\textbf{SemEval-Env}} & \multicolumn{2}{c}{\textbf{SemEval-Food}} & \multicolumn{2}{c}{\textbf{WordNet}} & \multicolumn{2}{c}{\textbf{Graphine}} \\
\cmidrule(lr){2-3} \cmidrule(lr){4-5} \cmidrule(lr){6-7} \cmidrule(lr){8-9} \cmidrule(lr){10-11}
\textbf{Metric} & \textbf{Acc} & \textbf{Wu\&P} & \textbf{Acc} & \textbf{Wu\&P} & \textbf{Acc} & \textbf{Wu\&P}  & \textbf{Acc} & \textbf{Wu\&P} & \textbf{Acc} & \textbf{Wu\&P} \\\midrule
\multicolumn{10}{l}{\emph{Self-supervised Setting}} \\
\textbf{TaxoExpan} & 27.8 & 57.6	& 11.1 & 54.8 & 27.6 & 54.2 & 19.8 & 64.8 & 24.5 & 65.9 \\ 
\textbf{STEAM} & 36.5 & 68.2 & 36.1 & 69.6 & 34.2 & 67.0 & 23.2 & 62.4 & 20.3 & 63.1 \\ 
\textbf{HEF}  &  53.6  &  75.6  &  55.3  &  71.4 &  47.9  & 73.5 & 16.4 & 60.3 & 25.5 & 66.5 \\
\textbf{Musubu} & 44.9 & 76.2 & 45.3 & 65.4 & 42.3 & 72.4 & 28.5 & 64.0 & 35.4 & \underline{75.2} \\
\textbf{TEMP} & 57.8 & 85.3 & 49.2 & 77.7 & 47.6 & 81.0 & 29.4 & 65.7 & \underline{35.9} & 73.8 \\
\textbf{BoxTaxo} & 31.8 & 64.7 & 38.1 & 75.4 & 31.4 & 66.8 & 26.4 & 63.9 & 29.2 & 68.2 \\
\textbf{TaxoPrompt}& \underline{61.4} & \underline{85.6} & \underline{57.4} & \underline{83.6} & \underline{53.2} & \underline{83.1} & 40.3 & 71.5 & 33.9 & 74.4 \\
\textbf{TaxoInstruct} & 45.9 & 76.2 & 48.8 & 77.2 & 34.3 & 70.2 & \underline{43.3} & \underline{71.8} & 31.8 & 69.0 \\ \midrule
\multicolumn{10}{l}{\emph{1-shot Setting}} \\
\textbf{NL} (GPT-4o) & 54.8 & \underline{88.3} & \underline{52.5} & \underline{81.3} & 55.5 & \underline{85.6} & \underline{72.2} & \underline{90.7} & \underline{69.8} & \underline{89.1} \\
\textsc{CodeTaxo} (GPT-4o) & \textbf{67.7} & \textbf{89.2} & \textbf{62.5} & \textbf{86.1} & \textbf{58.1} & 85.3 & \textbf{74.5} & \textbf{91.3} & \textbf{72.9} & \textbf{91.0}\\
\textbf{NL} (GPT-4o-mini) & 50.0 & 83.0 & 35.0 & 76.1 & 55.1 & \textbf{87.2} & 60.1  & 86.0 & 58.3 & 85.2\\
\textsc{CodeTaxo} (GPT-4o-mini) & \underline{58.1} & 85.6 & 42.5 & 76.0 & \underline{55.9} & 85.3 & 68.8 & 89.2 & 61.5 & 85.1\\ \midrule
\multicolumn{10}{l}{\emph{5-shot Setting}} \\
\textbf{NL} (GPT-4o) & 56.5 & 84.3 & \underline{60.0} & \underline{85.5} & 52.5 & 86.9 & \underline{72.2} & \underline{90.1} & 69.3 & \underline{90.0}\\
\textsc{CodeTaxo} (GPT-4o) & \textbf{66.1} & \textbf{88.0} & \textbf{67.5} & \textbf{87.0} & \textbf{60.2} & 85.7 & \textbf{76.5}  & \textbf{91.9} & \textbf{77.6} & \textbf{93.4}\\ 
\textbf{NL} (GPT-4o-mini)  & 53.2 & \underline{84.8} & 42.5 & 80.2 & 57.2 & \underline{87.6} & 63.4 & 87.3 & 63.5 & 88.6\\
\textsc{CodeTaxo} (GPT-4o-mini) & \underline{59.7} & \underline{84.8} & 47.5 & 78.3 & \underline{58.9} & \textbf{87.9} & 66.8 & 88.6 & \underline{70.3} & 89.1 \\
\bottomrule
\end{tabular} }
\caption{Performance on taxonomy expansion across two small-scale taxonomies (WordNet and Graphine) and three large-scale taxonomies (SemEval2016: science, environment, food). Bold indicates the highest score; underlined indicates the second-highest. All metrics are in percentages (\%).}
\label{tab:main_results}
\end{table*}

\subsubsection{Semantic Similarity Filter}
Including all entity $e \in \mathcal{E}_0$ to represent the existing taxonomy $\mathcal{T}_0$ presents two problems. First, large-scale taxonomies overload the LLM’s limited context window. Second, it unnecessarily expands the search space, introducing irrelevant entities and redundant information. To mitigate these issues, we propose a Semantic Similarity Filter that selects only entities relevant to the query $q$ for inclusion in the prompt context.

To compute the similarity between a query entity $q$ with its descriptive text $\mathcal{X}q = \{X^t_q, X^d_q\}$ and an entity $e_i \in \mathcal{E}0$ with its descriptive text $\mathcal{X}{e_i} = \{X^t_{e_i}, X^d_{e_i}\}$, we employ the pre-trained language model (PLM) as textual encoder. We concatenate the query entity $q$ and the $i$-th entity $e_i$ with special tokens [CLS] and [SEP], then encode the sequence using a pre-trained SimCSE model~\cite{gao2021simcse}. SimCSE converts them into $m$-dimensional representation $ \mathbf{q}, \mathbf{e}_i \in \mathbb{R}^m$:
\begin{align*}
    \mathbf{q} &= \textsc{PLM} (\textsc{[CLS]} \oplus X^t_q \oplus X^d_q  \oplus\textsc{[SEP]}) \\
    \mathbf{e}_i &= \textsc{PLM} (\textsc{[CLS]} \oplus X^t_{e_i} \oplus  X^d_{e_i} \oplus \textsc{[SEP]})
\end{align*}

The semantic relevance is calculated using cosine similarity between $\{\mathbf{e}_i\}^n_{i=1}$ and $\mathbf{q}$. We select the Top-$k$ entities most similar to query entity $q$ from the entity set $\mathcal{E}_0$ in $\mathcal{T}_0$ as follow: 
\begin{align*}
    \mathcal{I} &= \underset{\substack{\mathcal{I} \subseteq \{1, 2, \ldots, n\}, \\ |\mathcal{I}| = k}}{\operatorname{argmax}} \sum_{i \in \mathcal{I}} \text{cos\_sim}(\mathbf{e}_i, \mathbf{q})
\end{align*}
where $\mathcal{I}$ is the index set of the selected entities $\mathcal{E}_{sel} = \{e_i | i \in \mathcal{I}\}$ that represents the existing taxonomy. $k$ is set to 50\% of the entities in $\mathcal{E}_0$.

\subsubsection{Code Completion Prompt.}
\label{sec:code_completion_prompt}
The code completion prompt involves the instantiation of a query entity $q$ as an instance of the \texttt{Entity} class, as defined in Section~\ref{sec:entity_class_definition}. Since the query entity $q$ lacks information about its parent and child entities, it is instantiated as follows:
\begin{align*}
    q = \operatorname{Entity}(\text{name} &= {X}^{t}_q, \text{description} ={X}^{d}_q, \\ \text{parent} &= \text{None}, \text{child} = []) 
\end{align*}
Here, $X^t_q$ and $X^d_q$ define the query's name and description, while \texttt{None} and \texttt{[]} indicate the absence of parent and child entities.

We include the requirement ``\textit{Find the parent of the query node}'' as a comment to guide LLMs in selecting an anchor entity $a \in \mathcal{E}_{sel}$ as the parent entity for entity $q$. The output is the query $q$, an instance of the \texttt{Entity} class, which invokes the predefined method \texttt{add\_parent()} to assign $a$ as its parent entity like \texttt{q.add\_parent(a)}. 

We propose incorporating an optional feature in the code completion prompt: ``\textit{then generating a comment to explain why it is the parent of the given query node}''. This feature allows the LLM to simultaneously generate both the prediction and its rationale, improving explainability and revealing interesting insights, as discussed in Section~\ref{sec:case}.

\subsection{Few-shot Demonstration Construction}
To enhance LLMs' ability to expand our existing taxonomy, we propose a method for constructing demonstrations using the initial taxonomy $\mathcal{T}_0$. Our demonstration selection strategy focuses on the semantic similarity between the query entity $q$ and entities $e \in \mathcal{E}_0$ in the existing taxonomy. Specifically, we use SimCSE encoding to calculate these similarities, selecting the top-5 entities from the existing set $\mathcal{E}_0$ based on their similarity to $q$:
\begin{align*}
    \mathcal{I}_d &= \underset{\substack{\mathcal{I}_d \subseteq \{1, 2, \ldots, n\}, \\ |\mathcal{I}_d| = 5}}{\operatorname{argmax}} \sum_{i \in \mathcal{I}_d} \text{cos\_sim}(\mathbf{e}_i, \mathbf{q})
\end{align*}
Here, $\mathcal{I}_d$ represents the indices of entities selected for the demonstration set $\mathcal{E}_{demo} = \{e_i | i \in \mathcal{I}_d\}$. For each demonstration $d_i$, we treat each entity $e_i \in \mathcal{E}_{demo}$ as a query entity and, following the procedure outlined in Section~\ref{sec:code_completion_prompt}, add its parent entity using the \texttt{add\_parent} method.

\section{Experiments}

\subsection{Experimental Settings}
\paragraph{Datasets.} 
We evaluate taxonomy expansion on small-scale WordNet sub-taxonomies~\cite{bansal2014structured} and Graphine taxonomies~\cite{liu2021graphine}.  Additionally, we evaluate three large-scale taxonomies from SemEval-2016~\cite{bordea2016semeval} across science, environment, and food domains. For all benchmarks, 20\% of leaf entities are reserved for testing, with the remaining entities used for training. See App.~\ref{app:dataset} for details.

\paragraph{Baselines.} We evaluate \textsc{CodeTaxo}, using both \textsc{GPT-4o} and \textsc{GPT-4o-mini}, against self-supervised baselines including TaxoExpan~\cite{shen2020taxoexpan}, STEAM~\cite{yu2020steam}, HEF~\cite{wang2022qen}, Musubu~\cite{takeoka2021low}, TEMP~\cite{liu2021temp}, BoxTaxo~\cite{jiang2023single}, TaxoPrompt~\cite{xu2022taxoprompt}, and TaxoInstruct~\cite{shen2024unified}, and prompting LLMs through natural language. See details in App.~\ref{app:baseline}.

\begin{table*}[t]
\centering
\resizebox{1\textwidth}{!}{
\begin{tabular}{lccccccccccc}

\toprule
\multicolumn{1}{l}{\multirow{2}{*}{\makecell[c]{\textbf{Method}}}} & \multicolumn{1}{l}{\multirow{2}{*}{\makecell[c]{\textbf{Def.}}}} & \multicolumn{2}{c}{\textbf{SemEval-Sci}} & \multicolumn{2}{c}{\textbf{SemEval-Env}} & \multicolumn{2}{c}{\textbf{SemEval-Food}} & \multicolumn{2}{c}{\textbf{WordNet}} & \multicolumn{2}{c}{\textbf{Graphine}} \\ 
\cmidrule(lr){3-4} \cmidrule(lr){5-6} \cmidrule(lr){7-8} \cmidrule(lr){9-10} \cmidrule(lr){11-12}
& & \textbf{Acc} & \textbf{Wu\&P} & \textbf{Acc} & \textbf{Wu\&P} & \textbf{Acc} & \textbf{Wu\&P} & \textbf{Acc} & \textbf{Wu\&P} & \textbf{Acc} & \textbf{Wu\&P} \\ \hline

\multicolumn{10}{l}{\emph{1-shot Setting}} \\
\textbf{NL} (GPT-4o) & $\checkmark$ & 54.8 & \underline{88.3} & \underline{52.5} & \underline{81.3} & 55.5 & \underline{85.6} & \underline{72.2} & \underline{90.7} & \underline{69.8} & \underline{89.1} \\
& $\times$ & 59.7 & 89.0 & 57.5 & 82.8 & 56.4 & 87.0 & 68.1 & 89.1 & 68.8 & 90.1 \\

\textsc{CodeTaxo} (GPT-4o) & $\checkmark$ & \textbf{67.7} & \textbf{89.2} & \textbf{62.5} & \textbf{86.1} & \textbf{58.1} & 85.3 & \textbf{74.5} & \textbf{91.3} & \textbf{72.9} & \textbf{91.0} \\
& $\times$ & 56.5 & 84.5 & 55.0 & 85.1 & 56.8 & 86.1 & 66.4 & 88.4 & 69.8 & 88.8 \\ \midrule

\multicolumn{10}{l}{\emph{5-shot Setting}} \\
\textbf{NL} (GPT-4o) & $\checkmark$ & 56.5 & 84.3 & \underline{60.0} & \underline{85.5} & 52.5 & 86.9 & \underline{72.2} & \underline{90.1} & 69.3 & \underline{90.0} \\
& $\times$ & 59.7 & 89.6 & 50.0 & 79.3 & 55.5 & 87.6 & 70.5 & 89.9 & 68.8 & 88.9 \\

\textsc{CodeTaxo} (GPT-4o) & $\checkmark$ & \textbf{66.1} & \textbf{88.0} & \textbf{67.5} & \textbf{87.0} & \textbf{60.2} & 85.7 & \textbf{76.5} & \textbf{91.9} & \textbf{77.6} & \textbf{93.4} \\
& $\times$ & 51.6 & 80.6 & 65.0 & 86.7 & 57.6 & 86.1 & 67.8 & 88.8 & 68.8 & 89.7 \\

\bottomrule
\end{tabular} }
\vspace{-0.1in}
\caption{Impact of Definition Sentences (Def.) on \textsc{CodeTaxo} and NL Performance in 1-Shot and 5-Shot Settings.}
\vspace{-0.1in}
\label{tab:With/without_Definition}
\end{table*}

\paragraph{Evaluation Metrics:} We use two Accuracy (ACC) and Wu \& Palmer similarity (Wu\&P) to evaluate the performance of \textsc{CodeTaxo} and baselines. See details in App.~\ref{app:evaluation_metrics}

\begin{figure}[t]
    \centering
    \includegraphics[width=1\linewidth]{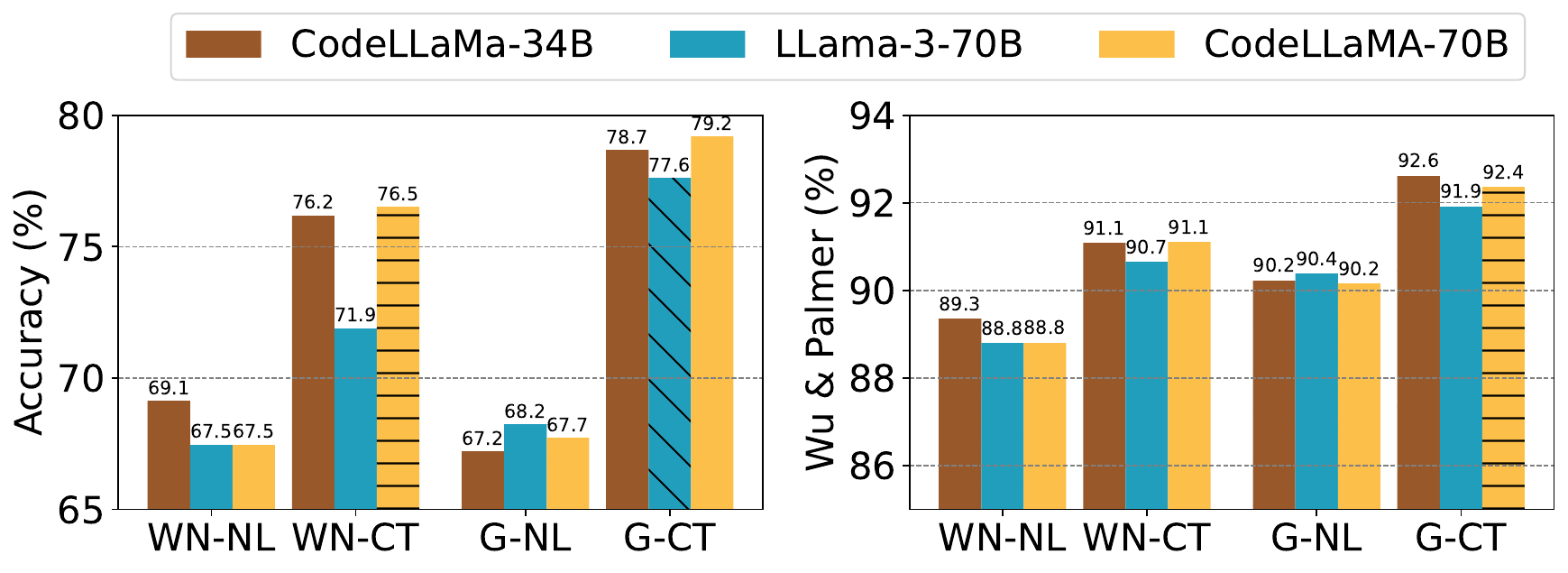}
    \vspace{-0.2in}
    \caption{Performance comparison of NL and \textsc{CodeTaxo} (CT) across Llama trained on Code and Natural Language domains. Due to limited contextual window sizes, evaluations were conducted on small-scale sub-taxonomies from WordNet (WN) and Graphine (G).}
    \vspace{-0.2in}
    \label{fig:CodeLLM}
\end{figure}

\subsection{Experimental Results}
\subsubsection{Can \textsc{CodeTaxo} expand taxonomy better than other baselines?}

We evaluate CodeTaxo against baseline methods for taxonomy expansion in Table~\ref{tab:main_results}, including self-supervised and in-context learning approaches. On WordNet and Graphine, both NL and \textsc{CodeTaxo} significantly outperform self-supervised baselines. In one-shot settings, \textsc{CodeTaxo} improves accuracy by 72.06\% and 103.06\% over the best self-supervised methods, demonstrating that minimal annotated data effectively unlocks LLMs' internal knowledge, while self-supervised methods struggle with limited-scale taxonomies. On large-scale SemEval-2016 taxonomies, \textsc{CodeTaxo} surpasses the best self-supervised baseline, TaxoPrompt, by 10.26\%, 8.89\%, and 9.21\% on SemEval-Sci, SemEval-Env, and SemEval-Food, respectively. While the NL prompt underperforms TaxoPrompt on SemEval-Sci and SemEval-Env, it exceeds TaxoPrompt on SemEval-Food but still trails CodeTaxo by 4.68\%, highlighting \textsc{CodeTaxo}’s superior ability to capture taxonomic structures. Performance depends on LLM capability and demonstration count, with GPT-4o outperforming GPT-4o-mini and more demonstrations improving accuracy across benchmarks, underscoring the value of high-quality demonstrations for taxonomy expansion.
\begin{table}[t]
\centering
\resizebox{1\columnwidth}{!}{
\begin{tabular}{lcccccccc}

\toprule
\multicolumn{1}{l}{\multirow{2}{*}{\makecell[c]{\textbf{Setting}}}} & \multicolumn{2}{c}{\makecell[c]{\textbf{Config.}}} & \multicolumn{2}{c}{\textbf{SemEval-Sci}} & \multicolumn{2}{c}{\textbf{SemEval-Env}} & \multicolumn{2}{c}{\textbf{SemEval-Food}} \\ 
\cmidrule(lr){2-3} \cmidrule(lr){4-5} \cmidrule(lr){6-7} \cmidrule(lr){8-9}
& \multicolumn{1}{c}{\textbf{Demo.}} & \multicolumn{1}{c}{\textbf{Filter}} & \textbf{Acc} & \textbf{Wu\&P} & \textbf{Acc} & \textbf{Wu\&P} & \textbf{Acc} & \textbf{Wu\&P}  \\ \hline

\multicolumn{1}{l}{\multirow{4}{*}{\makecell[c]{1-shot}}} & $\times$ & $\times$ & 50.0 & 84.0 & 47.5 & 81.1 & 56.4 & 85.3 \\
& $\times$ & $\checkmark$ & 61.3 & 84.4 & 55.0 & 83.2 & 54.2 & 84.6 \\
& $\checkmark$ & $\times$ & 61.3 & 85.9 & 47.5 & 79.0 & 57.2 & 86.7 \\
& $\checkmark$ & $\checkmark$ & 67.7 & 89.2 & 62.5 & 86.1 & 58.1 & 85.3 \\ \hline

\multicolumn{1}{l}{\multirow{4}{*}{\makecell[c]{5-shot}}} & $\times$ & $\times$ & 58.1 & 86.0 & 55.0 & 82.8 & 56.4 & 86.5 \\
& $\times$ & $\checkmark$ & 59.7 & 84.7 & 57.5 & 83.7 & 58.5 & 85.8 \\
& $\checkmark$ & $\times$ & 61.3 & 88.5 & 55.0 & 85.3 & 57.6 & 86.5 \\
& $\checkmark$ & $\checkmark$ & 66.1 & 88.0 & 67.5 & 87.02 & 60.2 & 85.7 \\

\bottomrule
\end{tabular} }
\caption{Ablation Study of two major modules in the \textsc{CodeTaxo}: All metrics are presented in percentages (\%). Configurations indicate whether Demonstration Selection (Demo.) and Semantic Similarity Filter (Filter) were employed.}
\vspace{-0.2in}
\label{tab:AblationStudy}
\end{table}

\subsubsection{How does \textsc{CodeTaxo} perform across different large language models?}

We evaluate \textsc{CodeTaxo}, a prompting method specifically designed for programming languages, by comparing its effectiveness against natural language prompting on both general-purpose LLMs and Code-LLMs. We used Llama-family models, including LLaMa-3-70B-instruct, CodeLLaMA-70B-instruct, and the smaller CodeLLaMA-34B-instruct, to evaluate how model size affects performance.  Given the limited contextual capacity of these models, we focused our evaluation on WordNet and Graphine, as shown in Figure~\ref{fig:CodeLLM}. The results highlight \textsc{CodeTaxo}'s superior accuracy and Wu\&P scores across all tested models, outperforming natural language prompts in representing taxonomic structures for black-box LLMs like GPT-4 and open-source LLMs. The analysis further reveals that \textsc{CodeTaxo} benefits more significantly from Code-LLMs, with a 13.33\% accuracy improvement on WordNet compared to 6.51\% for natural language prompts when transitioning to code language prompting. Notably, CodeLLaMA-34B-instruct, despite being smaller, showed better performance on WordNet and Graphine, emphasizing \textsc{CodeTaxo}'s efficiency and robustness.

\begin{figure}[t]
    \centering
    \includegraphics[width=1\linewidth]{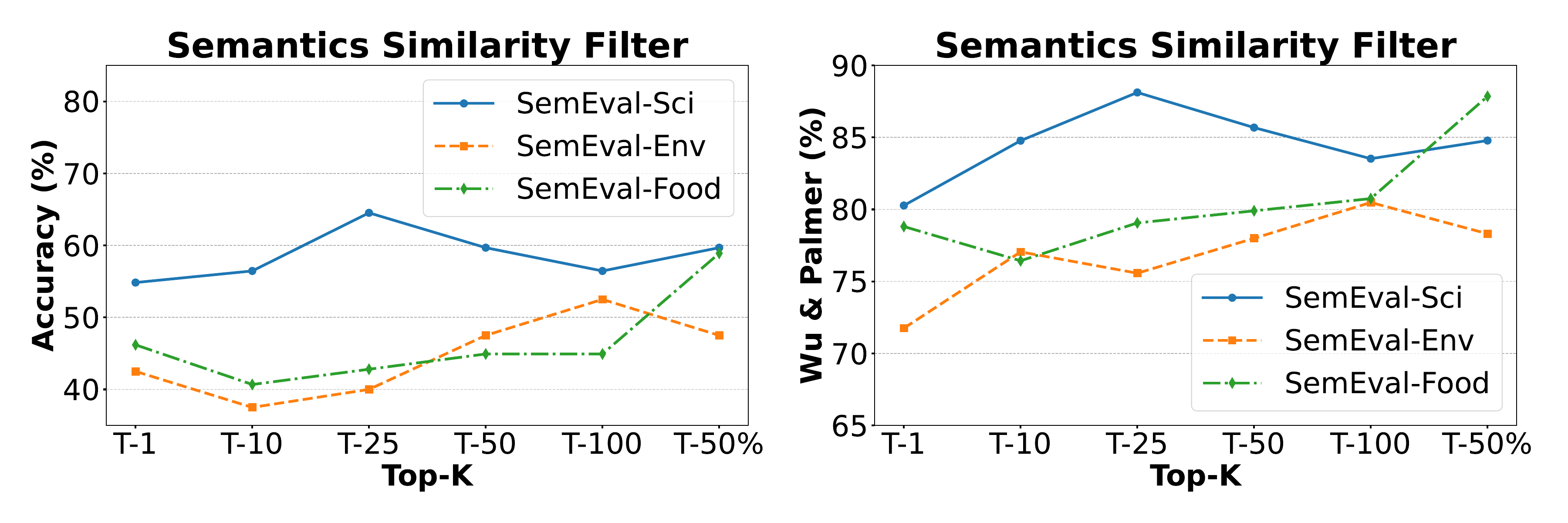}
    \vspace{-0.3in}
    \caption{Effect of Top-K relevant entities selected through SimCSE-based Semantic Similarity Filter.}
    \vspace{-0.2in}
    \label{fig:hit-at-n-performance}
\end{figure}


\subsection{Hyperparmeter Analysis of \textsc{CodeTaxo}}

This section explores the impact of selecting Top-K entities using the Semantic Similarity Filter on model performance, with experiments conducted using GPT-4o-mini across three SemEval taxonomies. As shown in Figure~\ref{fig:hit-at-n-performance}, increasing the number of Top-K entities generally enhances performance by retaining more entities, thereby reducing the likelihood of filtering out the ground truth and boosting prediction accuracy. However, this improvement involves a trade-off: a smaller search space sharpens the model's focus but increases the risk of excluding the ground truth. For instance, in the SemEval-Sci taxonomy, the model achieved optimal performance with a Hit@25 score of 78\% by retaining the top 25 entities, demonstrating the filter's ability to balance search space and coverage. To further refine this balance, we retained the top 50\% of entities in our experiments, ensuring that Hit@n exceeded 90\% across all benchmarks.

\begin{table*}[ht]
\small
\centering
\begin{tabular}{p{0.22\linewidth} p{0.22\linewidth} p{0.22\linewidth} p{0.22\linewidth}}
\toprule
\textbf{Query} & \textbf{NL Output} & \textbf{CodeTaxo Output} & \textbf{CodeTaxo Exp.}\\
\midrule
\textbf{functional genomics} \newline \textit{Def.} functional genomics is a field of molecular biology that attempts to describe \colorbox{yellow}{gene functions and} \colorbox{yellow}{interactions} & 

\textbf{genetics} 
\newline \textit{Def. }genetics is a branch of biology concerned with the study of \colorbox{green}{genes, genetic variation,} \colorbox{green}{and heredity} in organisms. &

\textbf{genomics} 
\newline \textit{Def. }genomics is an interdisciplinary field of biology focusing on the structure, function, evolution, mapping, and editing of genomes &
\textbf{genomics} 
\newline \textit{Exp. } Functional genomics is a subfield of genomics that focuses on understanding \colorbox{yellow}{the functions and inter-} \colorbox{yellow}{actions of genes} within the genome. \\
\midrule
\textbf{geostrategy} \newline \textit{Def. }geostrategy, a subfield of geopolitics, is a type of foreign policy guided principally by \colorbox{yellow}{geographical factors} as they inform, constrain, or affect political and military planning & 

\textbf{politics} \newline \textit{Def. }politics is the set of activities that are associated with making decisions in groups, or other forms of power relations between individuals, such as the distribution of resources or status & 
\textbf{geopolitics} 
\newline  \textit{Def. } geopolitics  on politics and international relations &
\textbf{geopolitics} 
\newline \textit{Exp. }: Geostrategy is a subfield of geopolitics, which focuses on \colorbox{yellow}{geographic factors} influencing political and military planning.\\
\bottomrule
\end{tabular}
\caption{Case study comparing the outputs of the \textsc{CodeTaxo} and NL prompt using the SemEval-Sci benchmarks. The table presents the definitions (Def.) of each model's prediction, additionally with the \textsc{CodeTaxo} explanations (Exp.) provided in the last column (CodeTaxo Exp.). Yellow highlights emphasize the specific focus of the query within its definition, as correctly captured by \textsc{CodeTaxo}, while green highlights indicate broader, less precise concepts used by the NL model.}
\label{table:case-study}
\vspace{-0.2in}
\end{table*}

\subsection{Ablation Study}
\subsubsection{Insight of Definition Sentences}
We performed an ablation study on definition sentences, a vital data source for taxonomy expansion tasks, using two prompting methods: NL and \textsc{CodeTaxo}. Our results in Table~\ref{tab:With/without_Definition} show that without definition sentences, \textsc{CodeTaxo} suffers a substantial drop in accuracy and Wu\&P across all benchmarks in both 1-shot and 5-shot settings, highlighting its reliance on semantic information from definitions to establish taxonomic relationships. Interestingly, NL performed better without definition sentences in specific benchmarks (SemEval-Sci, SemEval-Env, SemEval-Food) in the 1-shot setting, and in SemEval-Sci and SemEval-Food in the 5-shot setting. This suggests that NL struggles to process definition information effectively, potentially leading to incorrect predictions when overloaded with definitional content.

\subsubsection{Effevtiveness of Demo. Selection and Semantic Similarity Filter}
We performed an ablation study on the three SemEval2016 benchmarks mentioned above to assess the effectiveness of the two primary modules in \textsc{CodeTaxo}: Demonstration Selection (Demo.) and the Semantic Similarity Filter (Filter). Due to the relatively small size of the taxonomies in WordNet and Graphine, filtering redundant entities from the existing taxonomies was unnecessary. The results, presented in Table~\ref{tab:AblationStudy}, indicate that selecting demonstrations related to the query entity and filtering out unrelated entities in the existing taxonomy significantly improves taxonomy expansion. This finding suggests that incorporating more relevant contextual information and reducing redundant information to narrow the search space is beneficial for both accuracy and the Wu\&P score across all SemEval2016 benchmarks.

\subsection{Case Study}
\label{sec:case}
This section presents a case study demonstrating the effectiveness of our \textsc{CodeTaxo} framework by comparing its outputs to those of the natural language (NL) prompt, alongside model predictions and corresponding definitions in Table~\ref{table:case-study}. Notably, \textsc{CodeTaxo} aligns closely with the ground truth and generates explanations using the prompt from Section~\ref{sec:code_completion_prompt} to facilitate insightful discussions. For instance, in the query \textit{functional genomics}, \textsc{CodeTaxo} accurately classifies it under genomics, emphasizing its focus on the ``functions and interactions of genes within the genome'', whereas the NL model incorrectly selects the broader term \textit{genetics}. Similarly, \textsc{CodeTaxo} identifies \textit{geopolitics} as the parent entity of \textit{geostrategy}, highlighting its emphasis on \emph{geographic factors}, while the NL model selects the more general category of politics. These cases showcase \textsc{CodeTaxo}'s ability to leverage definition information for a comprehensive understanding of taxonomy structures, resulting in more precise predictions.


\section{Related Works}
\subsection{Taxonomy Expansion}
In taxonomy expansion, various approaches have been developed to integrate emerging entities into existing taxonomies. Aly \emph{et al.}~\cite{aly2019every} and Ma \emph{et al.} utilized hyperbolic embeddings to capture taxonomic relations, while Jiang \emph{et al.}~\cite{jiang2023single} and Xu \emph{et al.}~\cite{xu2024fuse} employed box embedding and fuse embedding instead of single vector embedding to encode taxonomic relations respectively. Manzoor \emph{et al.}~\cite{manzoor2020expanding} introduced implicit edge semantics to enhance entity representations. Self-supervised methods, such as Egonet~\cite{shen2020taxoexpan}, mini-path~\cite{yu2020steam}, and Ego-Tree~\cite{wang2021enquire}, have also been explored to model structural information within taxonomies.  To leverage more semantic information from the textural description of entities, Liu \emph{et al.}~\cite{liu2021temp}, Takeoka \emph{et al.}~\cite{takeoka2021low} and Xu \emph{et al.}~\cite{xu2022taxoprompt} fine-tuned BERT-based models to leverage textual descriptions of entities. Zhu \emph{et al.}~\cite{zhu2023towards} integrates textual and visual semantics to capture the hierarchical relation between entities. Shen \emph{et al.}~\cite{shen2024unified} and Moskvoretskii \emph{et al.}~\cite{moskvoretskii2024taxollama} unified framework combining various taxonomy construction tasks for instruction tuning. To our knowledge, \textsc{CodeTaxo} is the first work to perform taxonomy expansion via prompting LLMs.

\subsection{Code-LLMs for Structured Tasks}
Recent studies have demonstrated the strong performance of Code-LLMs in complex reasoning tasks~\cite{yang2024if, matraining}, including symbolic reasoning~\cite{madaan2022language, chengbinding}, graph reasoning~\cite{cai2024codegraph}, event structure prediction~\cite{wang2023code4struct,chen2023vistruct}, mathematical reasoning~\cite{gao2023pal}, and knowledge graph construction~\cite{li2023codeie,bi2024codekgc}. These works highlight Code-LLMs' ability to transform unstructured text into structured representations, enabling advanced reasoning tasks. In this paper, we focus on enhancing Code-LLMs' ability to comprehend and expand existing taxonomies through emerging query entities.

\section{Conclusion}
In this paper, we introduce \textsc{CodeTaxo}, a novel approach to taxonomy expansion that leverages code-based prompts to effectively utilize the inherent knowledge within LLMs. Our method addresses key challenges in taxonomy expansion by reformulating the task as a code completion problem and employing a Semantic Similarity Filtering mechanism to optimize the use of LLMs' contextual capacity. Extensive experiments on small-scale and large-scale taxonomies demonstrate that \textsc{CodeTaxo} achieves state-of-the-art performance in both one-shot settings and five-shot settings. We envision \textsc{CodeTaxo} as a powerful framework for integrating emerging entities into existing taxonomies by accurately identifying appropriate parent entities and also providing new insights for leveraging LLMs in structured knowledge tasks.

\section*{Limitations}
This study represents an initial effort to utilize LLMs for taxonomy expansion. Our primary objective is to identify an effective in-context learning strategy to leverage the potential of LLMs. We acknowledge that the performance and scalability of \textsc{CodeTaxo} are constrained by the inherent knowledge of LLMs and the limitations of their context window size. While this paper does not address the challenges of expanding LLM knowledge or increasing context window size, we hope that our work will inspire further research in these areas.

\section*{Ethics Statement}
Our research addresses taxonomy expansion within general knowledge domains, leveraging our proposed method, \textsc{CodeTaxo}, which uses large language models (LLMs) to generate structured knowledge and overcome the limitations of traditional manual taxonomy construction. We exclusively utilize publicly available datasets and benchmarks, avoiding user-generated, private, or sensitive data to ensure compliance with privacy and ethical standards. While our datasets do not engage directly with ethically sensitive content, LLMs inherently carry biases from their pre-training data, which may influence the structure and content of the expanded taxonomies. To address this, we integrate mechanisms for generating explanatory outputs, enabling detailed scrutiny of the model’s reasoning and identifying potential biases. Additionally, we recognize the risks of applying similar methodologies to subjective or sensitive domains, which could lead to misrepresentation or bias. To mitigate such risks, we emphasize collaboration with domain experts and advocate for responsible application of our methodologies across diverse fields, aiming to promote fairness, accuracy, and ethical research practices.


\newpage
\bibliography{anthology,custom,main}

\begin{thebibliography}{43}
\expandafter\ifx\csname natexlab\endcsname\relax\def\natexlab#1{#1}\fi

\bibitem[{Achiam et~al.(2023)Achiam, Adler, Agarwal, Ahmad, Akkaya, Aleman, Almeida, Altenschmidt, Altman, Anadkat et~al.}]{achiam2023gpt}
Josh Achiam, Steven Adler, Sandhini Agarwal, Lama Ahmad, Ilge Akkaya, Florencia~Leoni Aleman, Diogo Almeida, Janko Altenschmidt, Sam Altman, Shyamal Anadkat, et~al. 2023.
\newblock Gpt-4 technical report.
\newblock \emph{arXiv preprint arXiv:2303.08774}.

\bibitem[{Aly et~al.(2019)Aly, Acharya, Ossa, K{\"o}hn, Biemann, and Panchenko}]{aly2019every}
Rami Aly, Shantanu Acharya, Alexander Ossa, Arne K{\"o}hn, Chris Biemann, and Alexander Panchenko. 2019.
\newblock Every child should have parents: A taxonomy refinement algorithm based on hyperbolic term embeddings.
\newblock In \emph{Proceedings of the 57th Annual Meeting of the Association for Computational Linguistics}, pages 4811--4817.

\bibitem[{Bansal et~al.(2014)Bansal, Burkett, De~Melo, and Klein}]{bansal2014structured}
Mohit Bansal, David Burkett, Gerard De~Melo, and Dan Klein. 2014.
\newblock Structured learning for taxonomy induction with belief propagation.
\newblock In \emph{ACL}, pages 1041--1051.

\bibitem[{Bi et~al.(2024)Bi, Chen, Jiang, Xiong, Guo, Chen, and Zhang}]{bi2024codekgc}
Zhen Bi, Jing Chen, Yinuo Jiang, Feiyu Xiong, Wei Guo, Huajun Chen, and Ningyu Zhang. 2024.
\newblock Codekgc: Code language model for generative knowledge graph construction.
\newblock \emph{ACM Transactions on Asian and Low-Resource Language Information Processing}, 23(3):1--16.

\bibitem[{Bordea et~al.(2016)Bordea, Lefever, and Buitelaar}]{bordea2016semeval}
Georgeta Bordea, Els Lefever, and Paul Buitelaar. 2016.
\newblock Semeval-2016 task 13: Taxonomy extraction evaluation (texeval-2).
\newblock In \emph{Proceedings of the 10th international workshop on semantic evaluation (semeval-2016)}, pages 1081--1091.

\bibitem[{Cai et~al.(2024)Cai, Wang, Diao, Kwok, and Song}]{cai2024codegraph}
Qiaolong Cai, Zhaowei Wang, Shizhe Diao, James Kwok, and Yangqiu Song. 2024.
\newblock Codegraph: Enhancing graph reasoning of llms with code.
\newblock \emph{arXiv preprint arXiv:2408.13863}.

\bibitem[{Chen et~al.(2023)Chen, Wang, Li, Hoiem, and Ji}]{chen2023vistruct}
Yangyi Chen, Xingyao Wang, Manling Li, Derek Hoiem, and Heng Ji. 2023.
\newblock Vistruct: Visual structural knowledge extraction via curriculum guided code-vision representation.
\newblock In \emph{Proceedings of the 2023 Conference on Empirical Methods in Natural Language Processing}, pages 13342--13357.

\bibitem[{Cheng et~al.()Cheng, Xie, Shi, Li, Nadkarni, Hu, Xiong, Radev, Ostendorf, Zettlemoyer et~al.}]{chengbinding}
Zhoujun Cheng, Tianbao Xie, Peng Shi, Chengzu Li, Rahul Nadkarni, Yushi Hu, Caiming Xiong, Dragomir Radev, Mari Ostendorf, Luke Zettlemoyer, et~al.
\newblock Binding language models in symbolic languages.
\newblock In \emph{The Eleventh International Conference on Learning Representations}.

\bibitem[{Dubey et~al.(2024)Dubey, Jauhri, Pandey, Kadian, Al-Dahle, Letman, Mathur, Schelten, Yang, Fan et~al.}]{dubey2024llama}
Abhimanyu Dubey, Abhinav Jauhri, Abhinav Pandey, Abhishek Kadian, Ahmad Al-Dahle, Aiesha Letman, Akhil Mathur, Alan Schelten, Amy Yang, Angela Fan, et~al. 2024.
\newblock The llama 3 herd of models.
\newblock \emph{arXiv preprint arXiv:2407.21783}.

\bibitem[{Gao et~al.(2023)Gao, Madaan, Zhou, Alon, Liu, Yang, Callan, and Neubig}]{gao2023pal}
Luyu Gao, Aman Madaan, Shuyan Zhou, Uri Alon, Pengfei Liu, Yiming Yang, Jamie Callan, and Graham Neubig. 2023.
\newblock Pal: Program-aided language models.
\newblock In \emph{International Conference on Machine Learning}, pages 10764--10799. PMLR.

\bibitem[{Gao et~al.(2021)Gao, Yao, and Chen}]{gao2021simcse}
Tianyu Gao, Xingcheng Yao, and Danqi Chen. 2021.
\newblock Simcse: Simple contrastive learning of sentence embeddings.
\newblock In \emph{Proceedings of the 2021 Conference on Empirical Methods in Natural Language Processing}, pages 6894--6910.

\bibitem[{Huang et~al.(2019)Huang, Ren, Zhao, He, Wen, and Dong}]{huang2019taxonomy}
Jin Huang, Zhaochun Ren, Wayne~Xin Zhao, Gaole He, Ji-Rong Wen, and Daxiang Dong. 2019.
\newblock Taxonomy-aware multi-hop reasoning networks for sequential recommendation.
\newblock In \emph{Proceedings of the twelfth ACM international conference on web search and data mining}, pages 573--581.

\bibitem[{Jiang et~al.(2023)Jiang, Yao, Wang, and Sun}]{jiang2023single}
Song Jiang, Qiyue Yao, Qifan Wang, and Yizhou Sun. 2023.
\newblock A single vector is not enough: Taxonomy expansion via box embeddings.
\newblock In \emph{Proceedings of the ACM Web Conference 2023}, pages 2467--2476.

\bibitem[{Jurgens and Pilehvar(2016)}]{jurgens2016semeval}
David Jurgens and Mohammad~Taher Pilehvar. 2016.
\newblock Semeval-2016 task 14: Semantic taxonomy enrichment.
\newblock In \emph{Proceedings of the 10th international workshop on semantic evaluation (SemEval-2016)}, pages 1092--1102.

\bibitem[{Kang et~al.(2024)Kang, Agarwal, Jin, Lee, Yu, and Han}]{kang2024improving}
SeongKu Kang, Shivam Agarwal, Bowen Jin, Dongha Lee, Hwanjo Yu, and Jiawei Han. 2024.
\newblock Improving retrieval in theme-specific applications using a corpus topical taxonomy.
\newblock In \emph{Proceedings of the ACM on Web Conference 2024}, pages 1497--1508.

\bibitem[{Li et~al.(2023)Li, Sun, Tang, Yan, Wu, Huang, and Qiu}]{li2023codeie}
Peng Li, Tianxiang Sun, Qiong Tang, Hang Yan, Yuanbin Wu, Xuan-Jing Huang, and Xipeng Qiu. 2023.
\newblock Codeie: Large code generation models are better few-shot information extractors.
\newblock In \emph{Proceedings of the 61st Annual Meeting of the Association for Computational Linguistics (Volume 1: Long Papers)}, pages 15339--15353.

\bibitem[{Li et~al.(2024)Li, Zeng, Zuo, Ren, Liu, Su, Guo, Liu, Li, Hu et~al.}]{li2024knowcoder}
Zixuan Li, Yutao Zeng, Yuxin Zuo, Weicheng Ren, Wenxuan Liu, Miao Su, Yucan Guo, Yantao Liu, Xiang Li, Zhilei Hu, et~al. 2024.
\newblock Knowcoder: Coding structured knowledge into llms for universal information extraction.
\newblock \emph{arXiv preprint arXiv:2403.07969}.

\bibitem[{Liu et~al.(2020)Liu, Guo, Niu, Luo, Wang, Wen, and Xu}]{liu2020giant}
Bang Liu, Weidong Guo, Di~Niu, Jinwen Luo, Chaoyue Wang, Zhen Wen, and Yu~Xu. 2020.
\newblock Giant: scalable creation of a web-scale ontology.
\newblock In \emph{Proceedings of the 2020 ACM SIGMOD International Conference on Management of Data}, pages 393--409.

\bibitem[{Liu et~al.(2021{\natexlab{a}})Liu, Wang, Gu, Zhang, Zhang, and Wang}]{liu2021graphine}
Zequn Liu, Shukai Wang, Yiyang Gu, Ruiyi Zhang, Ming Zhang, and Sheng Wang. 2021{\natexlab{a}}.
\newblock Graphine: A dataset for graph-aware terminology definition generation.
\newblock In \emph{Proceedings of the 2021 Conference on Empirical Methods in Natural Language Processing}, pages 3453--3463.

\bibitem[{Liu et~al.(2021{\natexlab{b}})Liu, Xu, Wen, Jiang, Wu, and Yuan}]{liu2021temp}
Zichen Liu, Hongyuan Xu, Yanlong Wen, Ning Jiang, Haiying Wu, and Xiaojie Yuan. 2021{\natexlab{b}}.
\newblock Temp: Taxonomy expansion with dynamic margin loss through taxonomy-paths.
\newblock In \emph{Proceedings of the 2021 Conference on Empirical Methods in Natural Language Processing}, pages 3854--3863.

\bibitem[{MA et~al.()MA, Liu, Yu, Zhang, Jiang, Wang, and Li}]{matraining}
YINGWEI MA, Yue Liu, Yue Yu, Yuanliang Zhang, Yu~Jiang, Changjian Wang, and Shanshan Li.
\newblock At which training stage does code data help llms reasoning?
\newblock In \emph{The Twelfth International Conference on Learning Representations}.

\bibitem[{Madaan et~al.(2022)Madaan, Zhou, Alon, Yang, and Neubig}]{madaan2022language}
Aman Madaan, Shuyan Zhou, Uri Alon, Yiming Yang, and Graham Neubig. 2022.
\newblock Language models of code are few-shot commonsense learners.
\newblock In \emph{Proceedings of the 2022 Conference on Empirical Methods in Natural Language Processing}, pages 1384--1403.

\bibitem[{Manzoor et~al.(2020)Manzoor, Li, Shrouty, and Leskovec}]{manzoor2020expanding}
Emaad Manzoor, Rui Li, Dhananjay Shrouty, and Jure Leskovec. 2020.
\newblock Expanding taxonomies with implicit edge semantics.
\newblock In \emph{Proceedings of The Web Conference 2020}, pages 2044--2054.

\bibitem[{Moskvoretskii et~al.(2024)Moskvoretskii, Neminova, Lobanova, Panchenko, and Nikishina}]{moskvoretskii2024taxollama}
Viktor Moskvoretskii, Ekaterina Neminova, Alina Lobanova, Alexander Panchenko, and Irina Nikishina. 2024.
\newblock Taxollama: Wordnet-based model for solving multiple lexical sematic tasks.
\newblock \emph{arXiv preprint arXiv:2403.09207}.

\bibitem[{Shen et~al.(2020)Shen, Shen, Xiong, Wang, Wang, and Han}]{shen2020taxoexpan}
Jiaming Shen, Zhihong Shen, Chenyan Xiong, Chi Wang, Kuansan Wang, and Jiawei Han. 2020.
\newblock Taxoexpan: Self-supervised taxonomy expansion with position-enhanced graph neural network.
\newblock In \emph{Proceedings of The Web Conference 2020}, pages 486--497.

\bibitem[{Shen et~al.(2024)Shen, Zhang, Zhang, and Han}]{shen2024unified}
Yanzhen Shen, Yu~Zhang, Yunyi Zhang, and Jiawei Han. 2024.
\newblock A unified taxonomy-guided instruction tuning framework for entity set expansion and taxonomy expansion.
\newblock \emph{arXiv preprint arXiv:2402.13405}.

\bibitem[{Sun et~al.(2024{\natexlab{a}})Sun, Xu, Zha, Liu, and Dong}]{sun2024head}
Kai Sun, Yifan Xu, Hanwen Zha, Yue Liu, and Xin~Luna Dong. 2024{\natexlab{a}}.
\newblock Head-to-tail: How knowledgeable are large language models (llms)? aka will llms replace knowledge graphs?
\newblock In \emph{Proceedings of the 2024 Conference of the North American Chapter of the Association for Computational Linguistics: Human Language Technologies (Volume 1: Long Papers)}, pages 311--325.

\bibitem[{Sun et~al.(2024{\natexlab{b}})Sun, Xin, Sun, Xu, Yang, Dong, Tang, and Chen}]{sun2024large}
Yushi Sun, Hao Xin, Kai Sun, Yifan~Ethan Xu, Xiao Yang, Xin~Luna Dong, Nan Tang, and Lei Chen. 2024{\natexlab{b}}.
\newblock Are large language models a good replacement of taxonomies?
\newblock \emph{arXiv preprint arXiv:2406.11131}.

\bibitem[{Takeoka et~al.(2021)Takeoka, Akimoto, and Oyamada}]{takeoka2021low}
Kunihiro Takeoka, Kosuke Akimoto, and Masafumi Oyamada. 2021.
\newblock Low-resource taxonomy enrichment with pretrained language models.
\newblock In \emph{Proceedings of the 2021 Conference on Empirical Methods in Natural Language Processing}, pages 2747--2758.

\bibitem[{Tan et~al.(2022)Tan, Yang, Wei, Chen, Li, and Zheng}]{tan2022enhancing}
Yanchao Tan, Carl Yang, Xiangyu Wei, Chaochao Chen, Longfei Li, and Xiaolin Zheng. 2022.
\newblock Enhancing recommendation with automated tag taxonomy construction in hyperbolic space.
\newblock In \emph{2022 IEEE 38th International Conference on Data Engineering (ICDE)}, pages 1180--1192. IEEE.

\bibitem[{Touvron et~al.(2023)Touvron, Martin, Stone, Albert, Almahairi, Babaei, Bashlykov, Batra, Bhargava, Bhosale et~al.}]{touvron2023llama}
Hugo Touvron, Louis Martin, Kevin Stone, Peter Albert, Amjad Almahairi, Yasmine Babaei, Nikolay Bashlykov, Soumya Batra, Prajjwal Bhargava, Shruti Bhosale, et~al. 2023.
\newblock Llama 2: Open foundation and fine-tuned chat models.
\newblock \emph{arXiv preprint arXiv:2307.09288}.

\bibitem[{Wang et~al.(2021)Wang, Zhao, Chen, Zheng, and Liu}]{wang2021enquire}
Suyuchen Wang, Ruihui Zhao, Xi~Chen, Yefeng Zheng, and Bang Liu. 2021.
\newblock Enquire one’s parent and child before decision: Fully exploit hierarchical structure for self-supervised taxonomy expansion.
\newblock In \emph{Proceedings of the Web Conference 2021}, pages 3291--3304.

\bibitem[{Wang et~al.(2022)Wang, Zhao, Zheng, and Liu}]{wang2022qen}
Suyuchen Wang, Ruihui Zhao, Yefeng Zheng, and Bang Liu. 2022.
\newblock Qen: Applicable taxonomy completion via evaluating full taxonomic relations.
\newblock In \emph{Proceedings of the ACM Web Conference 2022}, pages 1008--1017.

\bibitem[{Wang et~al.(2023)Wang, Li, and Ji}]{wang2023code4struct}
Xingyao Wang, Sha Li, and Heng Ji. 2023.
\newblock Code4struct: Code generation for few-shot event structure prediction.
\newblock In \emph{Proceedings of the 61st Annual Meeting of the Association for Computational Linguistics (Volume 1: Long Papers)}, pages 3640--3663.

\bibitem[{Xu et~al.(2024)Xu, Jiang, Huang, Luo, Zhang, Chen, and Sun}]{xu2024fuse}
Fred Xu, Song Jiang, Zijie Huang, Xiao Luo, Shichang Zhang, Yuanzhou Chen, and Yizhou Sun. 2024.
\newblock Fuse: Measure-theoretic compact fuzzy set representation for taxonomy expansion.
\newblock In \emph{Findings of the Association for Computational Linguistics ACL 2024}, pages 2707--2720.

\bibitem[{Xu et~al.(2022)Xu, Chen, Liu, Wen, and Yuan}]{xu2022taxoprompt}
Hongyuan Xu, Yunong Chen, Zichen Liu, Yanlong Wen, and Xiaojie Yuan. 2022.
\newblock Taxoprompt: A prompt-based generation method with taxonomic context for self-supervised taxonomy expansion.
\newblock In \emph{IJCAI}, pages 4432--4438.

\bibitem[{Yang et~al.()Yang, Liu, Wu, Yang, Fung, Li, Huang, Cao, Wang, Ji et~al.}]{yang2024if}
Ke~Yang, Jiateng Liu, John Wu, Chaoqi Yang, Yi~Fung, Sha Li, Zixuan Huang, Xu~Cao, Xingyao Wang, Heng Ji, et~al.
\newblock If llm is the wizard, then code is the wand: A survey on how code empowers large language models to serve as intelligent agents.
\newblock In \emph{ICLR 2024 Workshop on Large Language Model (LLM) Agents}.

\bibitem[{Yang et~al.(2017)Yang, Zou, Wang, Yan, and Wen}]{yang2017efficiently}
Shuo Yang, Lei Zou, Zhongyuan Wang, Jun Yan, and Ji-Rong Wen. 2017.
\newblock Efficiently answering technical questions—a knowledge graph approach.
\newblock In \emph{Thirty-First AAAI Conference on Artificial Intelligence}.

\bibitem[{Ye et~al.(2022)Ye, Zhang, Chen, and Chen}]{ye2022generative}
Hongbin Ye, Ningyu Zhang, Hui Chen, and Huajun Chen. 2022.
\newblock Generative knowledge graph construction: A review.
\newblock In \emph{Proceedings of the 2022 Conference on Empirical Methods in Natural Language Processing}, pages 1--17.

\bibitem[{Yin and Shah(2010)}]{yin2010building}
Xiaoxin Yin and Sarthak Shah. 2010.
\newblock Building taxonomy of web search intents for name entity queries.
\newblock In \emph{Proceedings of the 19th international conference on World wide web}, pages 1001--1010.

\bibitem[{Yu et~al.(2020)Yu, Li, Shen, Feng, Sun, and Zhang}]{yu2020steam}
Yue Yu, Yinghao Li, Jiaming Shen, Hao Feng, Jimeng Sun, and Chao Zhang. 2020.
\newblock Steam: Self-supervised taxonomy expansion with mini-paths.
\newblock In \emph{Proceedings of the 26th ACM SIGKDD International Conference on Knowledge Discovery \& Data Mining}, pages 1026--1035.

\bibitem[{Zhang et~al.(2014)Zhang, Ahmed, Josifovski, and Smola}]{zhang2014taxonomy}
Yuchen Zhang, Amr Ahmed, Vanja Josifovski, and Alexander Smola. 2014.
\newblock Taxonomy discovery for personalized recommendation.
\newblock In \emph{Proceedings of the 7th ACM international conference on Web search and data mining}, pages 243--252.

\bibitem[{Zhu et~al.(2023)Zhu, Liu, Liang, Jiang, Xiao, Wang, Xie, and Xian}]{zhu2023towards}
Tinghui Zhu, Jingping Liu, Jiaqing Liang, Haiyun Jiang, Yanghua Xiao, Zongyu Wang, Rui Xie, and Yunsen Xian. 2023.
\newblock Towards visual taxonomy expansion.
\newblock In \emph{Proceedings of the 31st ACM International Conference on Multimedia}, pages 6481--6490.

\end{thebibliography}
\bibliographystyle{acl_natbib}

\appendix
\section{Appendix}
\label{sec:appendix}

\subsection{Datasets}
\label{app:dataset}

We evaluate the performance of taxonomy expansion methods on small-scale taxonomies using WordNet Sub-taxonomies from~\cite{bansal2014structured}, and Graphine taxonomies from~\cite{liu2021graphine}. Specifically, we use 35 Graphine taxonomies with fewer than 100 entities, selected from a total of 227 taxonomies. For the Graphine dataset, we selected 35 taxonomies with fewer than 100 entities out of 227 total taxonomies. In our experiment with WordNet, we utilized 114 sub-taxonomies from the test sets. Additionally, we evaluate three large-scale taxonomies from SemEval-2016~\cite{bordea2016semeval} across science, environment, and food domains. Table~\ref{tab:datasets} presents the statistics of these taxonomies, all of which contain entities and definitions curated by human experts. For all benchmarks, 20\% of leaf entities are reserved for testing, with the remaining entities used for training.

\begin{table}[h]
    \centering\Scale[0.8]{\begin{tabular}{lcccc}
    \toprule
    & \#Concepts & \#Edges & Depth & License\\
    \midrule
    WordNet & 20.5   & 19.5  & 3.0 &  WordNet\\ 
    Graphine &  48.2  &  48.2 & 4.6 & None\\
    SemEval-Sci & 429.0 &  451.0  &  8.0 & None\\
    SemEval-Env & 261.0 & 261.0  & 6.0 & None \\
    SemEval-Food & 1,486.0  & 1,576.0  &  8.0 & None \\
    \bottomrule
    \end{tabular}}
    \caption{Statistics of five taxonomy benchmarks. For WordNet and Graphine, we report the average for taxonomies included in these two benchmarks.}
    \label{tab:datasets}
\end{table}

\subsection{Baselines}
\label{app:baseline}

We compare our method with the following baselines for taxonomy expansion, all experiments are implemented in a server with three NVIDIA A6000 GPUs:
\begin{compactitem}
    \item \textbf{TaxoExpan~\cite{shen2020taxoexpan}:} adopts GNNs to encode local ego-graphs in taxonomy to enhance entity representation.
    \item \textbf{STEAM~\cite{yu2020steam}:} utilizes the mini-path information to capture the global structure of the taxonomy.
    \item \textbf{HEF~\cite{wang2022qen}:} represents taxonomies as ego-trees to capture hierarchy, fully leveraging the hierarchical structure to improve taxonomy coherence.
    \item \textbf{Musubu~\cite{takeoka2021low}:}  leverages pre-trained models and fine-tunes them as sentence classifiers using queries generated from Hearst patterns.
    \item \textbf{TEMP~\cite{liu2021temp}:}  utilizes a pre-trained model to encode text descriptions of each concept in the taxonomy. It incorporates taxonomic structure information through taxonomy paths.
    \item \textbf{BoxTaxo~\cite{jiang2023single}:} represent the entities via box embeddings instead of single vector embeddings to capture the hierarchical relation between entities.
    \item \textbf{TaxoPrompt~\cite{xu2022taxoprompt}:} adopt prompt tuning on the BERT-based encoder model to capture the taxonomic structure.
    \item \textbf{TaxoInstruct~\cite{shen2024unified}:} a unified framework for taxonomy-related tasks using instruction tuning, focused solely on taxonomy expansion for fair comparison.
\end{compactitem}
To the best of our knowledge, \textsc{CodeTaxo} represents the first work to address taxonomy expansion using an in-context learning approach. To validate the effectiveness of the code language based prompt design, we additionally propose a prompting method based on natural language prompts. The results obtained using the natural language prompt (NL) are presented in Table~\ref{tab:main_results}. To ensure that the natural language prompt communicates the same information as code language prompt in \textsc{CodeTaxo}, we represent each entity using natural language to describe its surface name, definition, parent, and children list.  The details of the NL prompt are provided in Table~\ref{tab:example}.  For a more direct comparison, we also demonstrate \textsc{CodeTaxo}'s predictions on the same example in Table~\ref{tab:codeexample}.

\subsection{Evaluation Metrics.} 
\label{app:evaluation_metrics}
The performance of \textsc{CodeTaxo} and the baseline models for taxonomy expansion tasks is evaluated using commonly adopted metrics, including accuracy (Acc) and Wu \& Palmer similarity (Wu\&P), as established in prior work~\cite{yu2020steam,liu2021temp, wang2021enquire}. Since \textsc{CodeTaxo} is a generation-based method rather than a ranking-based one, the mean reciprocal rank (MRR) used in the baselines is not applicable to \textsc{CodeTaxo}.

\begin{table}[t]
    \centering\Scale[0.75]{\begin{tabular}{lccccc}
    \toprule
    \multirow{2}{*}{\textbf{Dataset}} & \multicolumn{2}{c}{\textbf{1-shot}} & & \multicolumn{2}{c}{\textbf{5-shot}} \\
    \cmidrule(lr){2-3} \cmidrule(lr){5-6}
    & \textbf{NL} & \textbf{CodeTaxo} & & \textbf{NL} & \textbf{CodeTaxo} \\
    \midrule
    SemEval-Sci & 15737.2 & 9701.4 & & 16095.1 & 10342.6 \\ 
    SemEval-Env & 8965.7 & 5693.6 & & 9325.0 & 6321.1 \\
    SemEval-Food & 48908.1 & 30536.5 & & 49266.4 & 31176.3 \\
    WordNet & 948.9 & 1369.2 & & 1306.4 & 1962.3 \\
    Graphine & 2486.0 & 3223.9 & & 2855.2 & 3893.3 \\
    \bottomrule
    \end{tabular}}
    \caption{Comparison of average tokens used by NL and \textsc{CodeTaxo} across 5 benchmarks in 1-shot and 5-shot settings.}
    \label{tab:num-tokens}
\end{table}

\subsection{Efficiency Analysis of \textsc{CodeTaxo}}
\paragraph{Token Consumption}

Table~\ref{tab:num-tokens}  compares average token usage across benchmarks and prompt types (\textsc{CodeTaxo} vs. NL) in 1-shot and 5-shot settings. The findings highlight \textsc{CodeTaxo}'s efficiency in reducing token usage while maintaining effectiveness. Notably, in SemEval2016, \textsc{CodeTaxo} cuts token usage by approximately 37.6\% in the SemEval-Food task compared to natural language prompts. However, in the WordNet and Graphine datasets, \textsc{CodeTaxo} uses slightly more tokens due to the need to define Entity classes and methods. Overall, the significant reduction in token usage in SemEval2016 underscores \textsc{CodeTaxo}'s efficiency, especially in contexts with limited token windows.

\begin{table*}[t]
    \centering
    \resizebox{1\linewidth}{!}{
    \begin{tabularx}{\linewidth}{X}
        \toprule[1.5pt]
        \textbf{Natural Language Prompt} \\ \midrule[0.75pt]
        \textbf{User:} Given the current taxonomy, find the parent of the query node. Please note that the query node may be a new node not in the current taxonomy. The parent of given query node always exists, so do not generate 'none' or 'not found'. You only need to answer the entity name and do not generate any additional content or comments. \\
        \\
        lunacy: obsolete terms for legal insanity; parent: insanity; children: [].\\
        irrationality: the state of being irrational; lacking powers of understanding; parent: insanity; children: [].\\
        dementia: mental deterioration of organic or functional origin; parent: insanity; children: ['presenile dementia', 'alcoholic dementia', 'senile dementia'].\\
        alcoholic dementia: dementia observed during the last stages of severe chronic alcoholism; involves loss of memory for recent events although long term memory is intact; parent: dementia; children: [].\\
        Pick's disease: a progressive form of presenile dementia found most often in middle-aged and elderly women and characterized by degeneration of the frontal and temporal lobes with loss of intellectual ability and transitory aphasia; parent: presenile dementia; children: [].\\
        derangement: a state of mental disturbance and disorientation; parent: insanity; children: [].\\
        craziness: informal terms for insanity; parent: insanity; children: [].\\
        presenile dementia: dementia with onset before the age of 65; parent: dementia; children: ["Pick's disease"].\\
        senile dementia: dementia of the aged; results from degeneration of the brain in the absence of cerebrovascular disease; parent: dementia; children: [].\\
        insanity: relatively permanent disorder of the mind; parent: None; children: ['irrationality', 'dementia', 'craziness', 'derangement', 'lunacy'].\\
        \\
        Query node: Alzheimer's disease\\
        The parent of query node: \\
        \\
        \textbf{Assistant:} dementia \\
        \\
        \textbf{Ground Truth:} presenile dementia \\ \bottomrule[1.5pt]
        \end{tabularx}
    }
    \caption{Example of Natural Language (NL) Prompt.}
    \label{tab:example}
\end{table*}
\begin{table*}[]
    \centering
    \small
    \resizebox{1\linewidth}{!}{
    \begin{tabularx}{\linewidth}{X}
        \toprule[1.5pt]
        \textbf{Code Prompt} \\ \midrule[0.75pt]
        \textbf{User:} Complete the next line of code according to the comments and the given code snippet. You need to find the parent of the query node in the given current taxonomy and use the \texttt{add\_parent} function. The parent of given query node always exists in the given current taxonomy, so do NOT generate node that is NOT in the given current taxonomy. Note that you only need to complete the next ONE line of code, do not generate any additional content or comments. \\
        \\
\texttt{from typing import List}
\\\\
\texttt{class Entity:}

\quad\texttt{def \_\_init\_\_(self, name: str, description: str, parent: str, child: List['Entity']):}

\quad\quad\texttt{self.name = name}

\quad\quad\texttt{self.description = description}

\quad\quad\texttt{self.parent = parent}

\quad\quad\texttt{self.child = child}

\quad\texttt{def add\_parent(self, parent: 'Entity'):}

\quad\quad\texttt{self.parent = parent.name}

\quad\quad\texttt{parent.add\_child(self)}

\quad\texttt{def add\_child(self, child: 'Entity'):}

\quad\quad\texttt{self.child.append(child)}
\\\\
\texttt{\# Creating entities and establishing parent-child relationship}

\texttt{lunacy = Entity(name='lunacy', description='obsolete terms for legal insanity', parent=insanity, child=[])}

\texttt{irrationality = Entity(name='irrationality', description='the state of being irrational; lacking powers of understanding', parent=insanity, child=[])}

\texttt{dementia = Entity(name='dementia', description='mental deterioration of organic or functional origin', parent=insanity, child=['presenile dementia', 'alcoholic dementia', 'senile dementia'])}

\texttt{alcoholic\_dementia = Entity(name='alcoholic dementia', description='dementia observed during the last stages of severe chronic alcoholism; involves loss of memory for recent events although long term memory is intact', parent=dementia, child=[])}

\texttt{Pick's\_disease = Entity(name='Pick's disease', description='a progressive form of presenile dementia found most often in middle-aged and elderly women and characterized by degeneration of the frontal and temporal lobes with loss of intellectual ability and transitory aphasia', parent=presenile dementia, child=[])}

\texttt{derangement = Entity(name='derangement', description='a state of mental disturbance and disorientation', parent=insanity, child=[])}

\texttt{craziness = Entity(name='craziness', description='informal terms for insanity', parent=insanity, child=[])}

\texttt{presenile\_dementia = Entity(name='presenile dementia', description='dementia with onset before the age of 65', parent=dementia, child=["Pick's disease"])}

\texttt{senile\_dementia = Entity(name='senile dementia', description='dementia of the aged; results from degeneration of the brain in the absence of cerebrovascular disease', parent=dementia, child=[])}

\texttt{insanity = Entity(name='insanity', description='relatively permanent disorder of the mind', parent=None, child=['irrationality', 'dementia', 'craziness', 'derangement', 'lunacy'])}
\\\\
\texttt{\# creating query node}

\texttt{Alzheimer's\_disease = Entity(name='Alzheimer's disease', description='a progressive form of presenile dementia that is similar to senile dementia except that it usually starts in the 40s or 50s; first symptoms are impaired memory which is followed by impaired thought and speech and finally complete helplessness', parent=None, child=[])}
\\\\
\texttt{\# Finding the parent of query node} \\
        \\
        \textbf{Assistant:} Alzheimer's\_disease.add\_parent(presenile\_dementia) \\
        \\
        \textbf{Ground Truth:} presenile dementia \\ \bottomrule[1.5pt]
        \end{tabularx}
    }
    \caption{Example of Code-based Prompt.}
    \label{tab:codeexample}
\end{table*}


\end{document}